\documentclass[10pt,twocolumn,letterpaper]{article}

\usepackage{iccv}              % To produce the CAMERA-READY version
\usepackage{multirow}
\usepackage[dvipsnames]{xcolor}

\definecolor{iccvblue}{rgb}{0.21,0.49,0.74}
\usepackage[pagebackref,breaklinks,colorlinks,allcolors=iccvblue]{hyperref}

\DeclareRobustCommand\onedot{\futurelet\@let@token\@onedot}
\def\@onedot{\ifx\@let@token.\else.\null\fi\xspace}

\makeatletter
\DeclareRobustCommand\onedot{\futurelet\@let@token\@onedot}
\def\@onedot{\ifx\@let@token.\else.\null\fi\xspace}

\def\eg{\emph{e.g}\onedot}

\usepackage[utf8]{inputenc} % allow utf-8 input
\usepackage[T1]{fontenc}    % use 8-bit T1 fonts
\usepackage{hyperref}       % hyperlinks
\usepackage{url}            % simple URL typesetting
\usepackage{booktabs}       % professional-quality tables
\usepackage{amsfonts}       % blackboard math symbols
\usepackage{nicefrac}       % compact symbols for 1/2, etc.
\usepackage{microtype}      % microtypography
\usepackage{xcolor}         % colors
\usepackage{algorithm}
\usepackage{algpseudocode}

\usepackage[accsupp]{axessibility}

\def\paperID{3985} % *** Enter the Paper ID here
\def\confName{ICCV}
\def\confYear{2025}

\title{Web Artifact Attacks Disrupt Vision Language Models}

\author{Maan Qraitem, Piotr Teterwak, Kate Saenko, Bryan A. Plummer \\
  Boston University \\
  \texttt{\{mqraitem, piotrt, saenko, bplum\}@bu.edu}}

\begin{document}
\maketitle
\begin{abstract}

Vision-language models (VLMs) (\eg, CLIP, LLaVA) are trained on large-scale, lightly curated web datasets, leading them to learn unintended correlations between semantic concepts and unrelated visual signals. These associations degrade model accuracy by causing predictions to rely on incidental patterns rather than genuine visual understanding. Prior work has weaponized these correlations as an attack vector to manipulate model predictions, such as inserting a deceiving class text onto the image in a ``typographic'' attack. These attacks succeed due to VLMs' text-heavy bias—a result of captions that echo visible words rather than describing content. However, this attack has focused solely on text that matches the target class exactly, overlooking a broader range of correlations, including non-matching text and graphical symbols, which arise from the abundance of branding content in web-scale data. To address this gap, we introduce ``artifact-based'' attacks: a novel class of  manipulations that mislead models using both non-matching text and graphical elements. Unlike typographic attacks, these artifacts are not predefined, making them simultaneously harder to defend against and more challenging to find. We address this by framing artifact attacks as a search problem and demonstrate their effectiveness across five datasets, with some artifacts reinforcing each other to reach 100\% attack success rates. These attacks transfer across models with up to 90\% effectiveness, making it possible to attack unseen models. To defend against these attacks, we extend prior work's artifact aware prompting to the graphical setting.  We see a moderate  reduction of success rates of up to 15\% relative to standard prompts, suggesting a promising direction for enhancing model robustness. Code: \url{https://github.com/mqraitem/Web-Artifact-Attacks}
\end{abstract}

\section{Introduction}

\begin{figure}[t!]
    \centering
    \includegraphics[width= \linewidth]{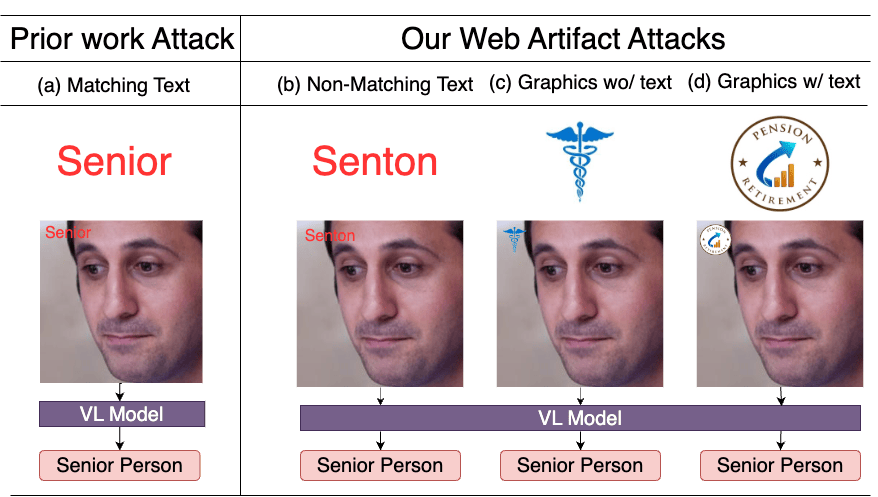}
    \vspace{-6mm}
    \caption{We propose a novel class of \textbf{Web Artifact Attacks} that expand the attack surface of Vision-Language Models beyond prior typographic attacks. For a sample task of predicting a person's age, (a) illustrates Typographic Attacks from prior work \citep{azuma2023defense} which manipulate predictions using class-matching text. In contrast, our proposed web artifact attacks, consisting of (b) non-matching text, (c) standalone graphics, and (d) graphics with embedded text, can also mislead the model.}
    \label{fig:intro_figure}
    \vspace{-4mm}
\end{figure}

\begin{figure*}[t!]
    \centering
    \includegraphics[width= 0.85\linewidth]{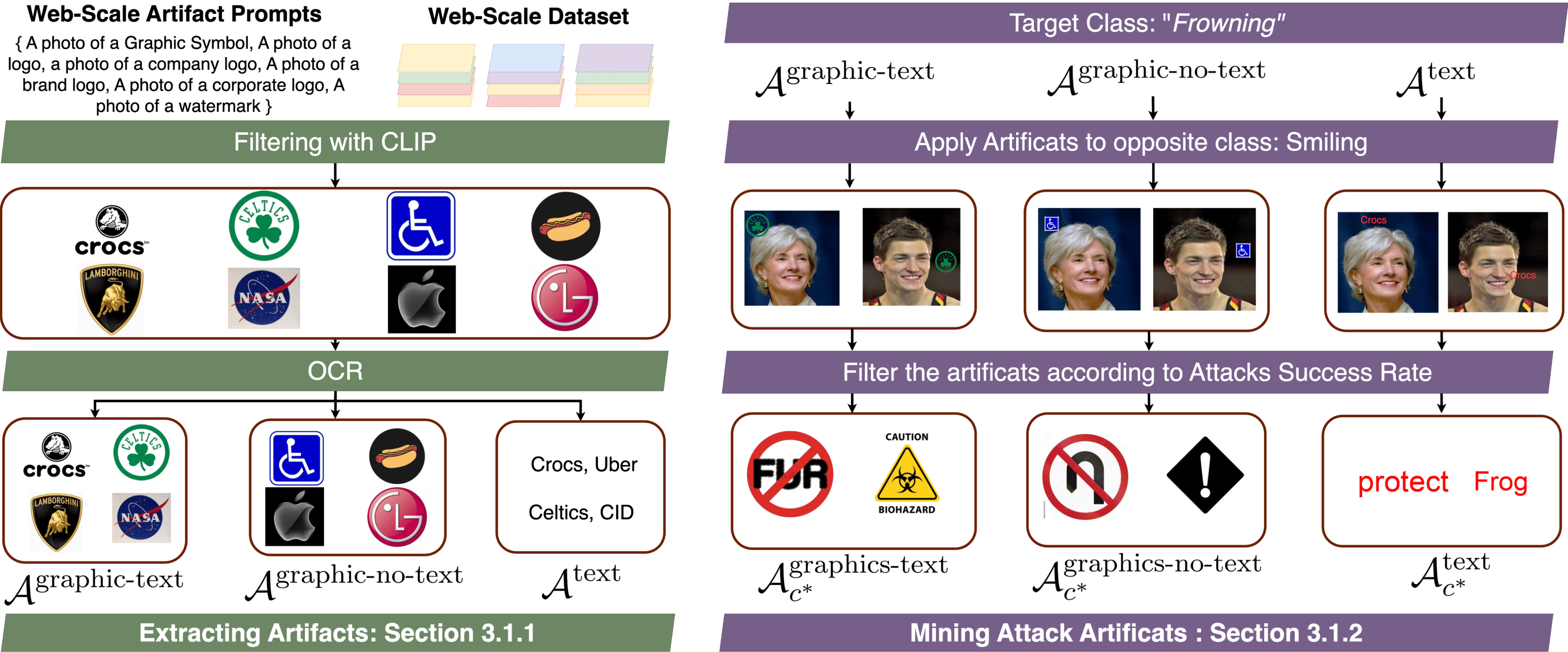}
    \vspace{-2mm}
    \caption{\textbf{Overview of the Web Artifacts Search and Selection}. The process involves two steps: \colorbox{ForestGreen}{\color{White}{Step 1:}} We filter an image-text dataset using CLIP-based retrieval and a set of artifact prompts. Retrieved artifacts are categorized into three groups: graphics with text ($\mathcal{A}^{\text{graphics-text}}$), graphics without text ($\mathcal{A}^{\text{graphics-no-text}}$), and unrelated text ($\mathcal{A}^{\text{text}}$) using OCR-based text extraction (\cref{subsec:artifact_extraction}). \colorbox{Purple}{\color{White}{Step 2:}} artifacts are applied to opposing-class images (e.g., "Smiling" images when targeting "Frowning"), and their effectiveness is evaluated based on their ability to mislead model predictions. Artifacts are ranked according to their attack success rate, resulting in a refined set of high-impact artifacts for each category ($\mathcal{A}_{c^*}^{\text{graphics-text}}$, $\mathcal{A}_{c^*}^{\text{graphics-no-text}}$, and $\mathcal{A}_{c^*}^{\text{text}}$) (\cref{subsec:artifact_mining}).} 
    \label{fig:method_figure}
    \vspace{-4mm}
\end{figure*}

Vision-language models (VLMs), such as CLIP \cite{radford2021learning} and LLaVA \cite{liu2023llava}, are trained on large-scale, lightly curated web datasets \cite{schuhmann2022laion,thomee2016yfcc100m}. Since web captions often describe what is present rather than what is important, models learn spurious patterns—unintended correlations that arise from frequent but unrelated co-occurrences \citep{zhou2022vlstereoset, agarwal2021evaluating,hall2023vision,janghorbani2023multimodal, wangsober,varma2024ravl,Neuhaus_2023_ICCV,li2023whac,lin2024parrot}.  For example, prior work found that Chinese characters frequently co-occur with the word “carton” in ImageNet \cite{li2023whac}, leading the model to misclassify images containing Chinese characters as a ``carton,'' regardless of its actual content. Moreover, these correlations can be weaponized by attackers to manipulate model predictions. \cref{fig:intro_figure}(a) shows a well-documented example, typographic attacks \citep{qraitem2024vision,azuma2023defense}, where inserting the text of the wrong category into an image tricks the model into predicting that category. This attack takes advantage of the model’s over-reliance on text, a bias reinforced by pretraining datasets in which captions frequently parrot text visible in the images \cite{lin2024parrot}. 

A shortcoming of prior work is that they focused only on explicit class-matching text, overlooking a broader set of vulnerabilities, including non-matching text and graphical artifacts such as logos. These elements, which are common in web-scale datasets due to branding and advertisements, may introduce unintended visual cues that models may rely on for classification. To address this gap, in \cref{fig:intro_figure}, we introduce Web Artifact Attacks, a broader class of attacks that exploit both non-matching text (\cref{fig:intro_figure}(b)) and graphics (\cref{fig:intro_figure}(c) and (d))  to mislead VLM predictions. We categorize graphical artifacts into two types: those with embedded text, which may reinforce textual biases, and those without text, isolating the impact of purely visual cues.

Unlike typographic attacks \citep{qraitem2024vision,azuma2023defense}, where the adversary inserts class-matching text to deliberately exploit the model’s reliance on words, Web Artifact Attacks pose a greater threat because the defender does not know in advance the attacking artifacts. Thus, it is not obvious how to identify effective attacking artifacts or defend against them. To find attacks, we frame Web Artifact Attacks as a search problem, scanning image-text datasets \cite{schuhmann2021laion, Changpinyo_2021_CVPR} to uncover text and graphics that can distort model predictions. The simplicity of these attacks makes them a notable security concern for deployed VLMs, as they only require dataset queries using similarity retrieval, making them highly accessible.

To that end, our search process consists of three stages: (1) \cref{fig:method_figure} \colorbox{ForestGreen}{\color{White}{Green}}: Artifact Retrieval, where we extract candidate artifacts by scanning image-text datasets for images with text or graphics (2) \cref{fig:method_figure} \colorbox{Purple}{\color{White}{Purple}}: Effect Estimation, where we apply artifacts to images from opposing classes and evaluate their influence on model predictions, and (3) placement optimization, where we refine the positioning of artifacts to maximize attack success. 

We evaluate the resulting attack artifacts across different VLM training objectives, visual encoder architectures, and pretraining dataset curation. Our findings demonstrate that Web Artifact Attacks are highly effective across various VLM training paradigms, including contrastive learning \cite{radford2021learning}, sigmoid loss \cite{zhai2023sigmoid}, and Large Language Model (LLM) fusion with visual features \cite{liu2023llava,zhu2023minigpt}. Furthermore, these vulnerabilities persist regardless of the visual encoder size (ViT-B-32, ViT-B-16, ViT-L-14), and attempts at stronger pretraining dataset curation (DataComp \cite{gadre2023datacomp}).

A key property of these attacks is high transferability—artifacts identified for one model retain up to 90\% effectiveness when applied to another, indicating shared failure modes across vision language  models. Additionally, attacks remain effective even when artifacts are small or transparent, revealing persistent vulnerabilities even in cases where artifacts are harder to detect. Furthermore, combining different artifact types significantly amplifies attack success, with text and graphical symbols reinforcing each other to reach nearly 100\% misclassification rates. While text-based artifacts and graphics with embedded text are the most effective, graphics without text provide a stealthier attack vector, making them harder to detect while still influencing predictions. 

Finally, we explore mitigation strategies by building on prior work \cite{cheng2024unveiling}, which showed that incorporating typographic attacks into the prompt can reduce their effectiveness. For example, if an image of a youth is attacked with the text ``senior,'' the prompt ``an image of a senior with the word `youth' written on it'' can weaken the attack. However, unlike typographic attacks, our attacks include graphical elements that cannot be easily incorporated into text. To address this, we propose augmenting the prompt with descriptions of graphical artifacts obtained through a captioning model. For LLM-based VLMs, we prompt the model to identify and describe these artifacts in its response. This approach reduces success rates by up to 15\% relative to standard prompts, suggesting a promising direction for enhancing model robustness.

To summarize our contributions: 

\begin{itemize}[nosep,leftmargin=*]
    \item We expose Web Artifact Attacks, which extend beyond typographic attacks by leveraging non-matching text, standalone graphics, and graphics with embedded text to mislead VLM predictions.  
    \item We develop a scalable attack pipeline to search, evaluate, and optimize artifacts from pretraining datasets, leveraging their natural abundance in branding and advertisements.  
    \item We show that our attacks are highly transferable, effective even when artifacts are small or transparent, and significantly stronger when combining multiple artifacts, reaching up to 100\% success rates. 
    \item We explore mitigation strategies that incorporate descriptions of the artifacts into the VLM(s) prompt which successfully mitigate some of their effect.
\end{itemize}

\vspace{-2mm}

\section{Related Work}

\noindent \textbf{Bias due to Artifacts.} Prior work \citep{li2023whac} has shown that models, from ResNet-50 \citep{He2016DeepRL} to large-scale VLMs \citep{radford2021learning,zhai2023sigmoid}, can mistakenly associate Chinese characters watermark with the carton class in Imagenet \cite{deng2009imagenet}. \citet{bykov2023mark} further demonstrated that multilingual text artifacts (\eg, Arabic, Latin, and Hindi) influence model predictions in unintended ways. Unlike these studies, which primarily observe such correlations, our work weaponizes these artifacts as an attack vector. Additionally, while prior work focused on manually discovered watermarks, we develop an automated mining pipeline that extracts a diverse set of non-matching text, graphical symbols with and without text at scale.

\noindent \textbf{Typographic Attacks}. Prior work \cite{Materzynska_2022_CVPR,azuma2023defense,qraitem2024vision,cheng2024unveiling} has demonstrated that VLMs, including CLIP \cite{radford2021learning} and LLaVA \cite{liu2023llava}, are highly susceptible to typographic attacks, where inserting class-matching text into an image directly manipulates predictions. These attacks exploit VLMs’ over-reliance on textual cues, a bias reinforced by pretraining datasets in which captions often echo the visible text within images \cite{lin2024parrot}. However, typographic attacks are inherently predefined, as they rely on explicitly matching class names, making them relatively easy to detect. In contrast, we show that non-matching text, graphical symbols, and mixed artifacts—elements not known a priori—can also mislead VLMs, enabling more flexible and covert attack strategies.

\noindent \textbf{Relation to Adversarial Attacks.} 
 Adversarial attacks 
\cite{yuan2021meta,rony2021augmented,zhu2021sparse,du2019query,chen2020universal} learn an imperceptible noise, that disrupts the model visual recognition capabilities when added to the model.  More recently, adversarial techniques have been weaponized for jailbreaking multimodal models to bypass safety mechanisms and generating restricted outputs \cite{qi2024visual,shayegani2023jailbreak,bagdasaryan2023abusing}. While our attacks share a similar end goal to adversarial attacks, namely, disrupting model behavior, adversarial and artifact attacks are fundamentally different from each other.  Our attacks originate as a result of spurious correlation in uncurated pretraining datasets, while adversarial attacks result from an artificial noise uniquely designed through the use of model intrinsics (\eg, gradients) to disrupt the fundamental mechanics of neural nets. Therefore, to develop such attacks, an attacker needs a sophisticated understanding of the intrinsic mechanics of neural nets such as \eg, Gradient  \cite{yuan2021meta,rony2021augmented,zhu2021sparse}, Meta Learning \cite{du2019query} or Attention maps \cite{chen2020universal} to name a few. However, the attacker, in our setup, only needs to know how to query the model.

\section{Web Artifact Attacks Against Vision-Language Models}
\label{sec:problem_def}

Vision-language models (VLMs) \cite{radford2021learning,liu2023llava,zhai2023sigmoid} are trained on large, uncurated web-scale datasets, leading them to form unintended correlations between visual elements and semantic categories \citep{wangsober,varma2024ravl,Neuhaus_2023_ICCV,li2023whac,lin2024parrot}. Prior work has demonstrated that typographic attacks, which insert explicit class-matching text into images, exploit these correlations to manipulate model predictions \citep{qraitem2024vision,azuma2023defense}. However, typographic attacks rely on direct textual matches, making them predictable and relatively easy to detect.

We propose Web Artifact Attacks, a new vector of unpredictable class of attacks that leverage artifacts from image-text datasets \cite{schuhmann2021laion, Changpinyo_2021_CVPR}. Unlike typographic attacks, these artifacts—such as non-matching text (text that contains partial overlaps, phonetic similarities, or visually similar characters but does not explicitly spell out the target class),  graphical symbols—do not need to explicitly match the target class yet can still mislead model predictions. 

Formally, our artifact-based attack introduces an artifact $a$ correlated with a target class $c^*$ into an image $x$, causing the model $f_\theta$ to misclassify it as $c^*$:
\begin{equation*}
\arg\max_{c} f_{\theta}(x) = \hat{c} \quad \Rightarrow \quad \arg\max_{c} f_{\theta}(x \oplus a) = c^*, 
\end{equation*}
\noindent where $\oplus$ denotes artifact insertion and $\hat{c}$ is the correct class.

Since these artifacts are not known a priori, they are harder to defend against while also making it challenging to determine in advance which artifacts will be effective. Therefore, we frame finding our attack artifacts as a search problem: systematically identifying artifacts within large-scale image-text datasets that can trigger misclassification. 

To that end, \cref{sec:attack_pipeline} describes our proposed the search pipeline, \cref{sec:art_insights} analyzes the extracted artifacts to reveal broader insights into model vulnerabilities, \cref{sec:success_rate_metrics} introduces an evaluation metric that accounts for occlusion effects,  and finally, \cref{sec:mitigation} proposes a mitigation strategy.

\subsection{Attack Pipeline}
\label{sec:attack_pipeline}

This section describes our search pipeline to systematically find attack artifacts. First, in \cref{subsec:artifact_extraction}, we extract artifacts from pretraining datasets by filtering images containing either text or graphical elements, categorizing them into non-matching text, graphics with text, and graphics without text. Next, in \cref{subsec:artifact_mining}, we evaluate how these artifacts influence model predictions by applying them to opposing-class images and selecting those that consistently increase misclassification rates. Finally, in \cref{subsec:artifact_placement}, we optimize artifact placement to maximize attack effectiveness.

\subsubsection{Extracting Artifacts from a Pretraining Dataset}
\label{subsec:artifact_extraction}

To search for our artifacts, we leverage the fact that graphics (\eg a logo of a company) frequently appear in web-scale datasets as standalone images due to the abundance of branding content. Using a CLIP-based retrieval mechanism, we apply broad text prompts $\mathcal{P}$ (\eg, “a photo of a graphic”) to extract images from an image-text dataset $\mathcal{D}$ (\eg, CC12M~\cite{Changpinyo_2021_CVPR}). We then rank the retrieved images by their similarity scores and retain the top 1\% of the highest-scoring images, ensuring that the selected artifacts closely match the intended query while filtering out irrelevant content. To verify the quality of the retrieved artifacts, we conduct a human study by randomly sampling 1,000 images from the extracted set and manually assessing their noise level, defined as the percentage of non-graphic or irrelevant images. We find that the noise level remains below 2\%, indicating that our retrieval process effectively isolates graphical artifacts while minimizing unrelated content.

Next, as shown in \cref{fig:method_figure}, we categorize artifacts into three types \colorbox{ForestGreen}{\color{White}{Green}} : unrelated text, graphics with embedded text, and graphics without text. To achieve this categorization, we apply OCR-based text detection (We use off-the-shelf models from EasyOCR \cite{EasyOCR}) to identify artifacts containing text and separate them from purely graphical artifacts. Additionally, to ensure that the extracted text does not merely replicate typographic attacks from prior work~\cite{azuma2023defense,qraitem2024vision}, we filter out any text that explicitly matches class names in the downstream dataset. This structured approach allows us to systematically evaluate how VLMs respond to textual and graphical cues, both independently and in combination, providing deeper insight into their learned vulnerabilities.

\subsubsection{Finding Artifacts That Influence a Target Class}
\label{subsec:artifact_mining}

After obtaining our set of artifacts in  \cref{subsec:artifact_extraction}, the next step is to determine which ones are most effective at misleading model predictions. Not all artifacts significantly influence classification—some may be ignored by the model, while others may reinforce correct predictions. To systematically identify high-impact artifacts, we evaluate their ability to shift model predictions toward a target class.

To that end, as \cref{fig:method_figure} \colorbox{Purple}{\color{White}{Purple}} illustrates, we apply each artifact to opposing-class images (\eg if the target class is ``car images''  then we use every image that is not a car) and measure its effectiveness in shifting predictions toward the target class If an artifact consistently causes images from unrelated classes to be classified as the target class, it reveals a strong learned correlation in the model that can be exploited for adversarial purposes \cite{qraitem2023bias}. Formally, given a downstream dataset $\mathcal{D}_{\text{downstream}}$, we define a subset of opposing-class images as:
\[
\mathcal{D}_{\neg c^*}^{\text{train}} \subset \{ x \mid x \in \mathcal{D}_{\text{downstream}}, y \neq c^* \},
\]
where $y$ is the ground-truth label. As we show in our experiments, $\mathcal{D}_{\neg c^*}^{\text{train}}$ can be as small as 10 samples per class while still uncovering the most effective artifacts.  We then apply each artifact $a$ from the three previously defined categories to this subset. 

% \[
% x_i^a = x_i \oplus a, \quad a \in \mathcal{A}_{\text{text}} \cup \mathcal{A}_{\text{graphics-text}} \cup \mathcal{A}_{\text{graphics-no-text}}
% \]
% where $\oplus$ denotes artifact insertion and $a$ is drawn from one of the three artifact categories: unrelated text ($\mathcal{A}_{\text{text}}$), graphics with embedded text ($\mathcal{A}_{\text{graphics-text}}$), or graphics without text ($\mathcal{A}_{\text{graphics-no-text}}$). 

To ensure efficient evaluation, artifacts are initially placed at random locations. This allows us to identify artifacts that demonstrate adversarial effects across a variety of placements, rather than optimizing for a single position prematurely. By deferring location-specific optimization, we reduce the computational complexity from $\mathcal{O}(|\mathcal{A}||\mathcal{D}_{\neg c^*}^{\text{train}}||\mathcal{L}|)$ to $\mathcal{O}(|\mathcal{A}||\mathcal{D}_{\neg c^*}^{\text{train}}|)$ during this stage (where $\mathcal{L}$ refers to the set of target locations), ensuring that we focus on selecting artifacts with strong generalizable effects before fine-tuning their positioning.

Artifact impact score $a$ is defined per-artifact as the proportion of modified images classified as the target class:
\[
P_{c^*}(a) = \frac{1}{n} \sum_{i=1}^{n} \mathbf{1}[f_\theta(x_i^a) = c^*], 
\]
where higher values indicate a stronger influence on misclassification. We compute this score for each artifact in our three categories: text artifacts ($\mathcal{A}^{\text{text}}$), graphics with embedded text ($\mathcal{A}^{\text{graphics-text}}$), and graphics without text ($\mathcal{A}^{\text{graphics-no-text}}$), as defined in \cref{subsec:artifact_extraction}. For each artifact category, we rank artifacts by $P_{c^*}(a)$ in descending order and select the top $s$ most effective ones. This results in the sets $\mathcal{A}_{c^*}^{\text{text}}$, $\mathcal{A}_{c^*}^{\text{graphics-text}}$, and $\mathcal{A}_{c^*}^{\text{graphics-no-text}}$, where $c^*$ denotes the target class.

\subsubsection{Optimizing Artifact Placement}
\label{subsec:artifact_placement}

While  \cref{subsec:artifact_mining} identified artifacts that are consistently effective across different placements, their positioning within an image can further enhance misclassification rates. 

To address this, for each artifact $a \in \mathcal{A}_{c^*}^{\text{top-text}} \cup \mathcal{A}_{c^*}^{\text{top-graphics-text}} \cup \mathcal{A}_{c^*}^{\text{top-graphics-no-text}}$, we evaluate a set of candidate locations $\mathcal{L}$ to determine the most effective positioning.  For each artifact, we pick the location that obtains the highest success rate. This step fine-tunes the effect of already strong artifacts, maximizing misclassification rates. Refer to the Supp.\ B for an ablation of location effect. In short, we find that the top border of the image (especially top middle) to be the most effective. 
% \[
% l^*(a) = \arg\max_{l \in \mathcal{L}} P_{c^*}(a, l)
% \]

\subsection{Insights from the Artifacts}  
\label{sec:art_insights}

\begin{figure*}[t!]
    \centering
    \includegraphics[width= 0.9\linewidth]
    {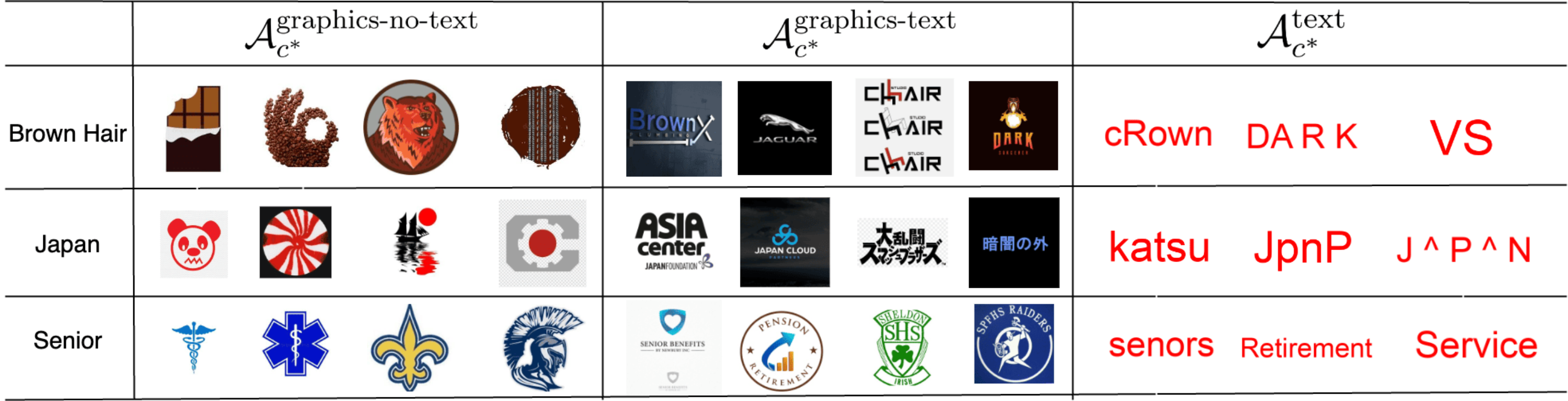}
    \vspace{-2mm}
    \caption{\textbf{Examples of Web Artifacts} split into: graphics without text (\( A^{\text{graphics-no-text}}_{c^*} \)), graphics with embedded text (\( A^{\text{graphics-text}}_{c^*} \)), and unrelated text (\( A^{\text{text}}_{c^*} \)). Each row corresponds to a different target class (\textit{Brown Hair}, \textit{Japan}, \textit{Adult}), that models have learned to associate with the artifacts. Notably, text artifacts need not match the class exactly, while graphical symbols can represent indirect but learned associations. These findings highlight the diverse range of artifacts that can manipulate model predictions. Refer to Supp.\ A for more examples.}
    \label{fig:examples_figure}
    \vspace{-4mm}
\end{figure*}

After extracting and ranking artifacts based on their influence on model predictions, we examine the highest-impact artifacts for different target classes. \cref{fig:examples_figure} presents examples of these artifacts, categorized into \textbf{graphics without text} ($\mathcal{A}_{c^*}^{\text{graphics-no-text}}$), \textbf{graphics with embedded text} ($\mathcal{A}_{c^*}^{\text{graphics-text}}$), and \textbf{unrelated text artifacts} ($\mathcal{A}_{c^*}^{\text{text}}$). These examples reveal distinct patterns in how different artifact types induce misclassification.  

\noindent \textbf{Non-textual graphics} often share  visual attributes loosely related to the target class. For instance, brown-colored logos and animal symbols appear frequently for ``Brown Hair,''  while ``Japan''  is associated with red-and-white designs reminiscent of its national flag. Similarly, the ``Senior''  class is linked to medical symbols and retirement-related imagery, suggesting that models rely on broad visual associations learned from pretraining data.  

\noindent \textbf{Graphics with embedded text} frequently contain contextually related words, even when the complete phrase is irrelevant. Examples include “Brown” from unrelated brand names appearing in the ``Brown Hair''  category and “Japan Cloud” for the ``Japan''  category. This demonstrates that models are highly sensitive to the presence of text fragments within graphical elements, further reinforcing the text-heavy biases introduced during pretraining.  

\noindent \textbf{Text-only artifacts}, we observe frequent misspellings, phonetic approximations, and visually similar character patterns rather than exact class matches. Words like “senors” for “Senior” or “J\textasciicircum P\textasciicircum N” for “Japan” illustrate how models generalize text-based associations beyond strict class labels, making these artifacts more difficult to detect than conventional typographic attacks.  

Refer to Supplementary A for more examples.

\subsection{Attack Success Evaluation}
\label{sec:success_rate_metrics}

Prior evaluations of typographic attacks \citep{Materzynska_2022_CVPR,azuma2023defense,qraitem2024vision,cheng2024unveiling} measure attack success by checking whether inserting class-matching text misleads VLMs. However, they overlook occlusion effects—misclassification may occur due to the added text obscuring important image features. To ensure that attack success is solely due to the artifact rather than occlusion, we introduce an evaluation framework that isolates its effect while preserving the image’s classifiability.

Given an input image $x$ and an artifact $a$ placed at location $l$, we define the attack-modified image as $x^a = x \oplus a$, where $\oplus$ denotes artifact insertion. To verify that misclassification stems from the artifact itself, we introduce a masking-based validation step. We generate a masked version of the image, $x^{\neg l}$, where the region $l$ is occluded using a mask $M_l$.

To ensure that the image remains classifiable after artifact removal, we retain only samples where $x^{\neg l}$ is still correctly classified as its original class $c$. Let $f_\theta$ be the model’s classifier, and let $c$ be the ground-truth class of $x$. We discard any image where $f_\theta(x^{\neg l}) \neq c$, ensuring that attack success cannot be attributed to the removal of critical visual features. 

\begin{figure*}[t!]
    \centering
    \includegraphics[width=0.95\linewidth]{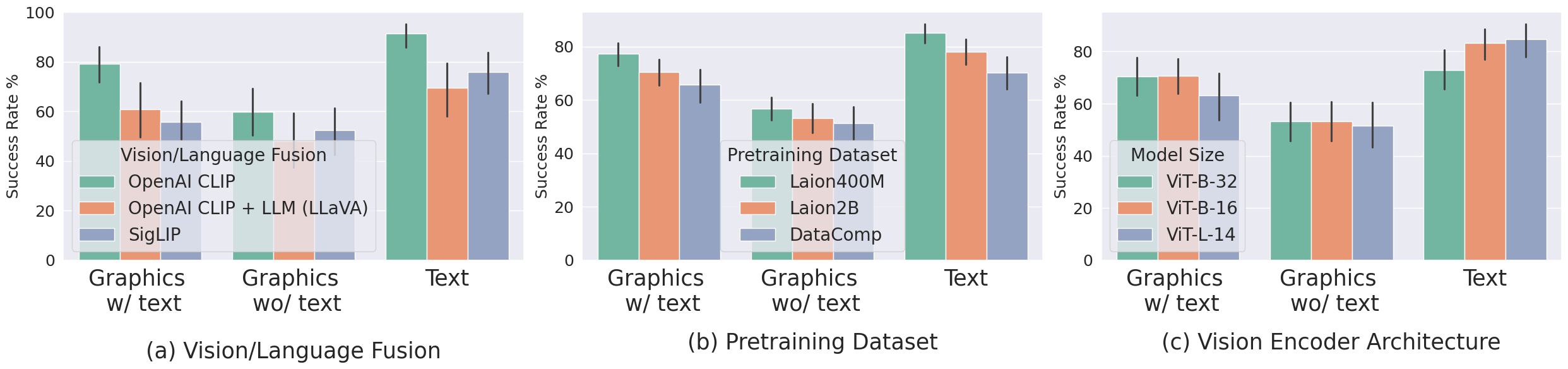}
    \vspace{-2mm}
    \caption{\textbf{Breakdown of Artifact Attacks Performance.} Our Web Artifact attacks are highly effective across different configurations: namely (a) Vision/Language Fusion, (b) Pretraining Dataset, and (c) Vision Encoder Architecture. Refer to \cref{sec:main_results} for further discussion..}
    \vspace{-4mm}
\label{fig:pretrained_vs_models_success}
\end{figure*}

The Attack Success Rate (ASR) is then defined as the proportion of cases where the model predicts the target class $c^*$ for $x^a$ but maintains correct classification for $x^{\neg l}$:
\[
ASR = \frac{1}{N} \sum_{i=1}^{N} \mathbf{1} \left[ f_\theta(x_i^a) = c^* \wedge f_\theta(x_i^{\neg l}) = c \right],
\]
where $\mathbf{1}[\cdot]$ is the indicator function, and $N$ is the number of samples passing the masking validation step. This stricter evaluation ensures that attack success is genuinely due to the artifact rather than incidental occlusion effects. To ensure that the attack generalizes beyond the images used during search, the final ASR is evaluated on a separate test set disjoint from the training set $\mathcal{D}_{\neg c^*}^{\text{train}}$ used for artifact mining.

\section{Experiments}
\label{sec:exps}

\noindent\textbf{Datasets.} We use CC12M \cite{Changpinyo_2021_CVPR} as our primary dataset for searching and extracting artifacts. Our extraction process yields 87k artifacts, but this approach is dataset-agnostic and can be applied to any large-scale image-text dataset, such as LAION \cite{schuhmann2021laion, schuhmann2022laion}. To evaluate the effectiveness of these artifacts, we conduct experiments across five distinct tasks spanning both human-related attributes and object/geography classification. For human-related tasks, we predict Gender (Male/Female) and Age (Child, Teen, Adult, Elderly) using FairFace \cite{karkkainen2021fairface}, as well as Smiling using CelebA \cite{LLWT15CelebA}. For non-human classification tasks, we predict aircraft models using FGVC Aircraft \cite{maji2013fine} and countries using Country211 \cite{radford2021learning}.  For each downstream dataset, we use only 32 images per class to estimate the Artifact Success Rate (ASR), which as we demonstrate in \cref{sec:main_results}, is sufficient to estimate ASR. 

\noindent\textbf{Metrics.} We use the success rate as outlined in \cref{sec:success_rate_metrics}. In short, given a target class $c^*$, we measure attack effectiveness by evaluating whether the model misclassifies an image after the artifact that correlates with  $c^*$ is introduced while ensuring that the image remains correctly classified when the artifact is masked. We also report the change of the confidence score in target class $c^*$ after pasting the artifact. 

\noindent\textbf{Models.} We evaluate the robustness of VLMs to Web Artifact Attacks by varying vision-language fusion strategy, vision encoder architecture, and pretraining dataset size and curation. For fusion, we test contrastive learning (CLIP) \cite{radford2021learning}, sigmoid-based alignment (SigLIP) \cite{zhai2023sigmoid}, and LLM-based fusion (LLaVA) \cite{liu2023llava}. To assess the impact of architecture size, we evaluate ViT-B-32, ViT-B-16, ViT-L-14 while we fix the pretraining dataset to LAION 2B \cite{schuhmann2021laion}. Finally, we compare models pretrained on LAION 400M, LAION-2B \cite{schuhmann2021laion}, and DataComp \cite{gadre2023datacomp} while fixing the pretraining architecture to ViT-B-32 and ViT-B-16 to examine whether dataset curation affects attack vulnerability.

\noindent\textbf{Artifact Setting.} We fix the artifact size to 10th of the image size. Refer to Supplementary. D for a visual example and ablation of artifact size and transparency. 

\subsection{Results}
\label{sec:main_results}

We benchmark the effect of Web Artifact Attacks on Vision-Language Models (VLMs) across different vision-language fusion methods, model architectures, and pretraining datasets. We make the following observations:   

\noindent\textbf{Artifact Attacks are Effective across Diverse Model Configurations.} As seen in \cref{fig:pretrained_vs_models_success}, Web Artifact Attacks consistently achieve high success rates across different vision-language fusion strategies, pretraining datasets, and model architectures, reaching up to 90\% success. Comparing the artifacts, text-based attacks are the most effective, followed by graphics with embedded text, and finally graphics without text, suggesting that models are most vulnerable to explicit textual cues but still susceptible to purely visual artifacts.

\noindent\textbf{CLIP is most vulnerable to Artifact Attacks.} In \cref{fig:pretrained_vs_models_success}(a), models trained with standard contrastive learning (OpenAI CLIP) exhibit the highest attack success rates across all artifact types, particularly for text-based artifacts, reinforcing CLIP’s strong reliance on textual cues. The LLM-based fusion model (LLaVA) shows a moderate reduction in vulnerability, likely due to its alignment on a smaller, curated dataset such as COCO \cite{lin2014microsoft}, which likely have much fewer graphics and logos. Similarly, SigLIP, which employs a different training objective and a pretraining dataset, achieves moderate reduction in vulnerability too on par with LLaVA.

\noindent\textbf{Pretraining Dataset Curation has Minimal Effect on Success Rate.} \cref{fig:pretrained_vs_models_success}(b) shows that models pretrained DataComp \cite{gadre2023datacomp} exhibit a slightly improved robusteness to artifact attacks than uncurated Laion400M and LAION 2B.  Despite DataComp’s filtering, it does not majorly reduce susceptibility, suggesting that its filtering strategy may be ill-suited for mitigating spurious correlations. DataComp primarily optimizes for image quality and semantic similarity to high-quality benchmarks, rather than explicitly removing textual artifacts or graphical elements that reinforce spurious correlations. As a result, while the dataset curation improves image-text alignment for downstream tasks, it fails to eliminate the kind of incidental patterns that make models vulnerable to Web Artifact Attacks. This highlights a critical limitation: filtering strategies that focus on dataset relevance may not address Web Artifact Attacks.

\noindent\textbf{Larger Architecture is More Vulnerable to Textual Attacks.} As seen in \cref{fig:pretrained_vs_models_success}(c), larger models, such as ViT-L-14, exhibit higher attack success rates on text-based attacks. This can be attributed to their increased capacity for optical character recognition, enabling them to more effectively detect and leverage textual elements in images. As a result, they become more susceptible to attacks that exploit text-based spurious correlations present in web-scale datasets.

\begin{figure}[t!]
    \centering
    \includegraphics[width= 1.0\linewidth]{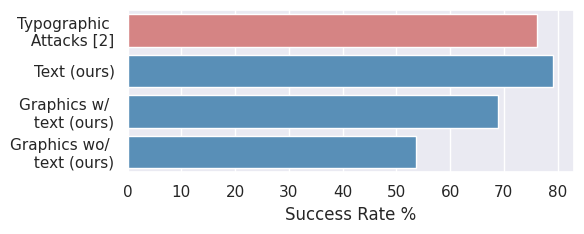}
    \vspace{-8mm}
    \caption{\textbf{Comparison to Prior Work Attacks}. Success rate of Web Artifact attacks vs prior work Typographic Attacks \cite{azuma2023defense}. }
    \label{fig:avg_results}
    \vspace{-2mm}
\end{figure}

\begin{figure}[t!]
    \centering
    \includegraphics[width= 0.95\linewidth]{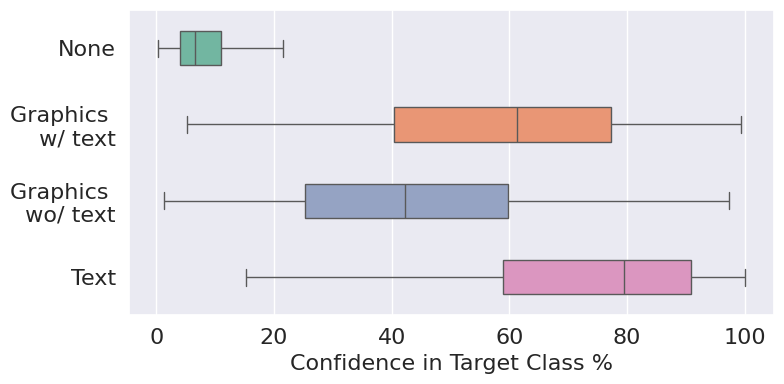}
    \vspace{-2mm}
    \caption{\textbf{Confidence in Target Class} when different types of artifacts—graphics with text (orange), graphics without text (green), and text (red)—are introduced, compared to the baseline with no artifacts (blue). Results demonstrate that artifacts significantly increase model confidence in the target class.}
    \label{fig:pretrained_vs_models_logits}
    \vspace{-2mm}
\end{figure}

\begin{figure}[t!]
    \centering
    \includegraphics[width= 0.9 \linewidth]{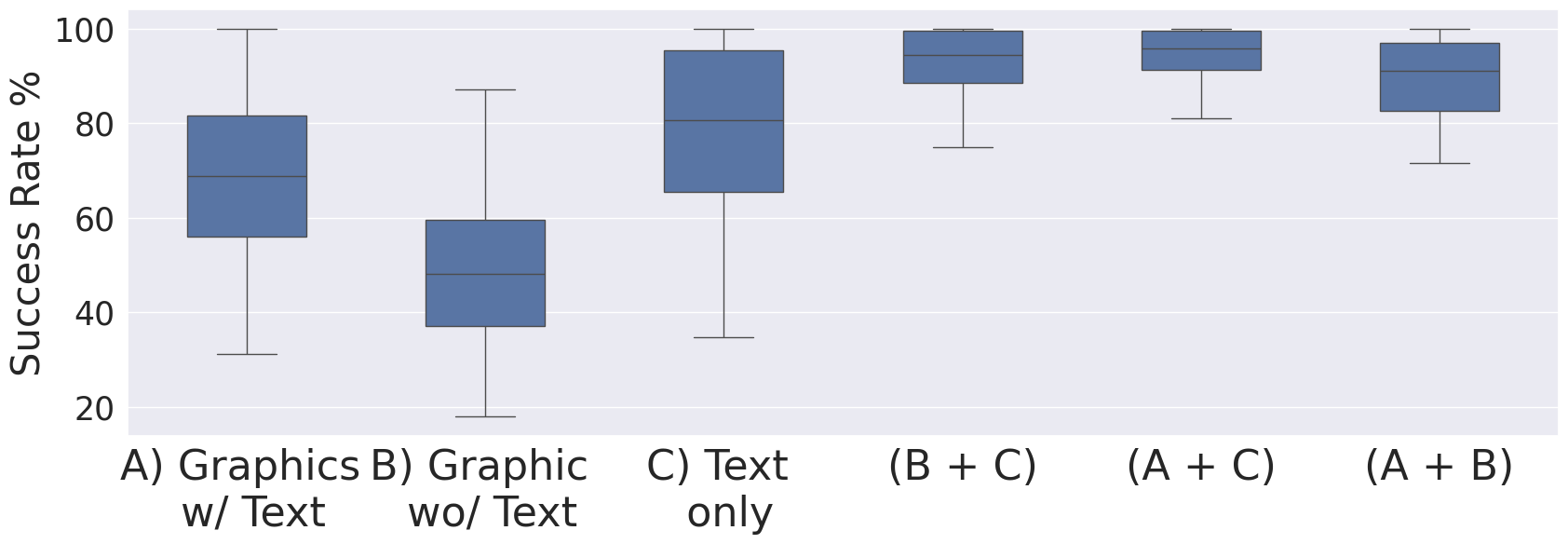}
    \vspace{-2mm}
    \caption{\textbf{Success Rate of Combining artifacts}. Results indicate that combining  artifact types leads to higher attack success.}
    \label{fig:artifacts_combined}
    \vspace{-2mm}
\end{figure}

\begin{figure}[t!]
    \centering
    \includegraphics[width= 0.85\linewidth]{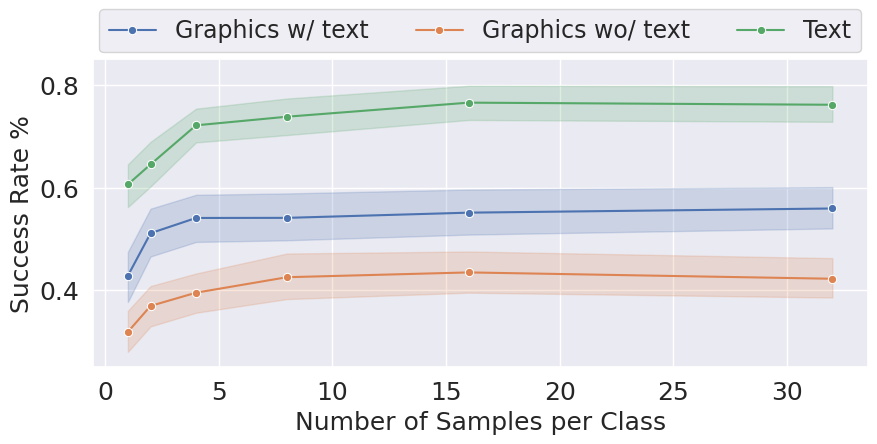}
    \vspace{-2mm}
    \caption{\textbf{Sample Size vs Success Rate}. Success rate of artifact-based attacks vs the number of samples used from the downstream dataset to estimate their effect. Results indicate that a relatively small sample size is sufficient to assess attack effectiveness.}
    \label{fig:num_subjects}
    \vspace{-6mm}
\end{figure}

\noindent\textbf{Artifacts Based Attacks vs Prior Work's Typographic Attacks.} \cref{fig:avg_results} presents the average performance of our web-scale Web Artifact Attacks compared to Typographic Attacks \cite{azuma2023defense}. The results demonstrate that our text-based attacks, derived from mined logo datasets, are more effective than typographic attacks, which are typically limited to using text from the opposing class. This suggests that optimizing attacks beyond simple class-matching text—by leveraging a broader range of textual configurations—can significantly enhance their effectiveness. In contrast, graphical attacks tend to be less effective overall. However, they open up a novel attack surface that, as discussed in Section \ref{sec:mitigation}, proves more challenging to defend against.

\noindent\textbf{Artifacts Increase Model Confidence in the Target Class.} \cref{fig:pretrained_vs_models_logits} shows that the introduction of textual and graphical artifacts significantly increases the model’s confidence in the target class compared to the no-artifact baseline. Text-based artifacts lead to the highest confidence scores, followed by graphics with embedded text and graphics without text. This increased confidence correlates with the attack success rates observed in prior experiments, suggesting that artifacts not only influence predictions but also enhance model certainty in incorrect classifications.

\noindent\textbf{The Compounding Effect of Different Types of Artifacts} \cref{fig:artifacts_combined} presents the attack success rate across different artifact types and their combinations. The results highlight a clear synergistic effect when multiple artifacts are used together, significantly increasing misclassification rates compared to individual artifacts. While text-only artifacts remain highly effective, their impact is amplified when combined with graphical symbols. The highest success rates are observed in (A + C) and (A + B) cases, where text is paired with either graphical symbols with or without text, reaching near 100\% attack success in some instances. This suggests that models do not treat textual and graphical cues independently but rather reinforce their reliance on both when they co-occur, exacerbating spurious correlations learned during pretraining. Additionally, graphical artifacts without text (B) alone exhibit moderate attack success, but when combined with text (B + C), their effectiveness increases substantially.

\begin{figure*}[t!]
    \centering
    \includegraphics[width= 0.9\linewidth]{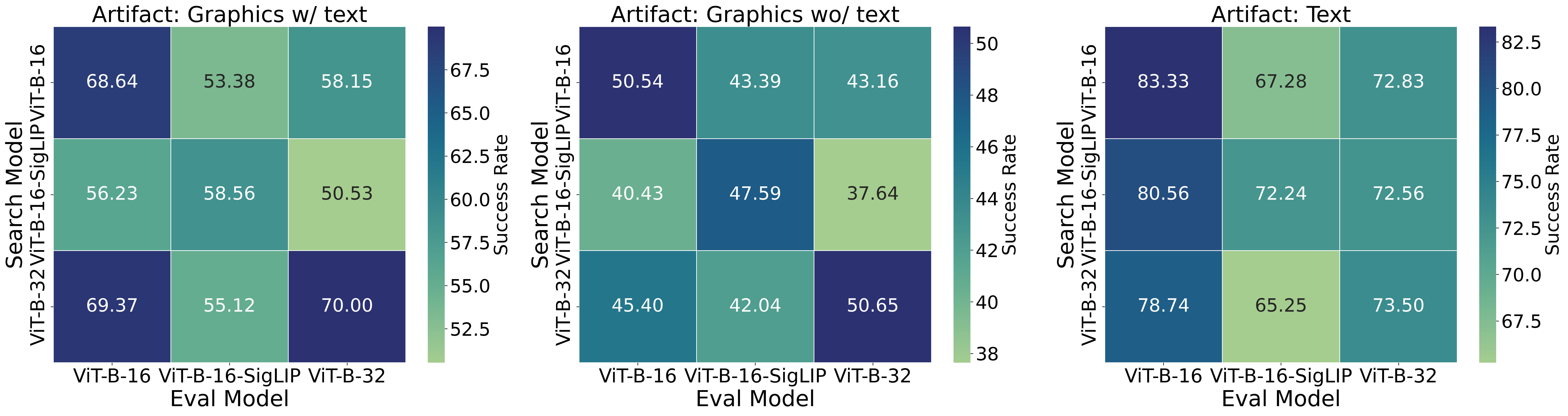}
    \vspace{-2mm}
    \caption{\textbf{Transferability of Web Artifact Attacks Across Different Vision Encoder Architectures.} The heatmap shows the success rate (\%) of artifacts search on a specific model (y-axis) and evaluated on another model (x-axis). Higher success rates indicate greater transferability of spurious correlations. Results indicate that artifacts effect is highly transferable across architectures.}
    \label{fig:heatmap_transfer}
    \vspace{-4mm}
\end{figure*}

\begin{figure}[t!]
    \centering
    \includegraphics[width=\linewidth]{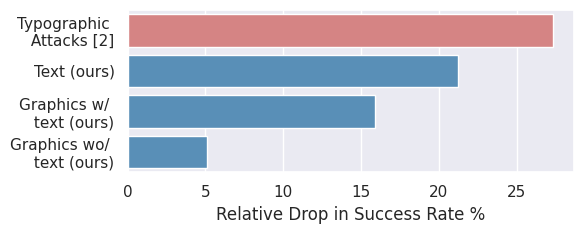}
    \vspace{-6mm}
    \caption{\textbf{Artifact Aware Prompting}. Results indicate that the attack effect can be mitigated with more informative prompts. }
    \label{fig:method_results}
    \vspace{-6mm}
\end{figure}

\noindent\textbf{A Small Sample Size is Enough for Estimating Artifact Effect}.  A key question for our attacks  is how many samples are needed per class to reliably estimate an artifact's attack effect. \cref{fig:num_subjects} shows that across all artifact types, the attack success rate stabilizes quickly as the number of samples increases, indicating that only a small sample size is required to obtain a reliable estimate. This result is significant as it suggests that evaluating Web Artifact Attacks does not require extensive sampling, which greatly reduces computational overhead. By using only a fraction of the dataset, we can efficiently assess the impact of artifacts while preserving computational resources. This finding is particularly important for large-scale evaluations, where repeatedly testing artifacts on full datasets would be prohibitively expensive.

\noindent\textbf{Web Artifact Attacks are Highly Transferable}. \cref{fig:heatmap_transfer} illustrates that Web Artifact Attacks are highly transferable across different models. Across all artifact types, success rates remain consistently high when applying artifacts mined from one model to another. The most significant drop in success rate occurs when transferring attacks to a model trained with SigLIP \cite{zhai2023sigmoid}, which employs a different training objective. However, even in this case, the attack success rates remain substantial, indicating that spurious correlations exploited by these attacks persist across diverse model architectures and training paradigms. This suggests that mitigating these vulnerabilities will require more than simply altering the training objective.

\section{Mitigation w/ Artifact Aware Prompting}
\label{sec:mitigation}

Recent work~\cite{cheng2024unveiling} has uncovered an important insight: VLMs can be guided to adjust their attention toward specific visual elements when provided with more informative prompts. This finding was leveraged as a mitigation strategy for typographic attacks by explicitly incorporating the attack text into the prompt. For example, assume the an image of a dog is attacked with the text of ``a cat,''  then rather than using ``a photo of a dog,'' the authors use: ``a photo of a dog with cat written on it.'' While this approach is effective for textual artifacts, it does not generalize to graphical artifacts, which lack an explicit textual form and cannot be directly incorporated into the language input.

To address this limitation, we propose a mitigation approach inspired by ~\cite{cheng2024unveiling} for the two types of models. For \textbf{Contrastive VLM(s)}: we use structured descriptions of graphical symbols to make these artifacts explicit to the model. Specifically, we  employ a captioning model to transform the graphics into a textual description which then are appended to the input prompt, ensuring that the model processes them explicitly. Refer to Supp.\ E for an example. For \textbf{Large VLM(s)}: following~\cite{cheng2024unveiling}, we leverage the generative capabilities for LVLMs like LLaVA \cite{liu2023llava} to integrate artifact awareness directly. We prompt LLaVA to describe any graphical or textual artifacts present in the image, ensuring that these features are explicitly considered during inference. Note that our approach does not require model retraining or architectural modifications. The only overhead is the cost of executing the captioning model.

\smallskip
\noindent\textbf{Results}. \cref{fig:method_results} illustrates the effectiveness of our approach in reducing attack success rates. Across all artifact types, artifact-aware prompting consistently lowers the success rate, indicating that explicitly guiding the model to consider artifacts can diminish their impact. However, this defense is less effective against our attacks compared to prior typographic attacks \cite{azuma2023defense}, with the largest performance gap observed for graphical attacks. This highlights the increased difficulty posed by our attacks and suggests promising directions for future research in developing more robust defenses.
\section{Conclusion}
\label{sec:conclusion}

In this work, we introduced \textbf{Web Artifact Attacks}, a novel vector of attacks that make use of non-matching text and graphical artifacts—both with and without embedded text— to mislead model predictions, revealing vulnerabilities beyond traditional typographic attacks. Our results show that these attacks not only increase model confidence in incorrect classifications but also persist across different model architectures, vision-language fusion strategies, and pretraining datasets. Moreover, we showed how our search process for attack artifacts is efficient only requiring a handful of samples to approximate attack effect. While mitigation strategies such as artifact-aware prompting reduce attack success rates, they are insufficient to eliminate the threat, particularly for non-textual graphical artifacts.

\noindent\textbf{Future Work.} Our findings highlight the need for more robust defenses against our attacks. A promising direction is dataset curation that extends beyond simple CLIP-based filtering; developing more informative captioning strategies that better contextualize images containing logos and graphics could help models learn more meaningful associations rather than relying on spurious correlations. Additionally, while our artifact discovery process is effective, it could be further optimized for efficiency. Future improvements could explore faster, more scalable methods for identifying artifacts in large-scale datasets, reducing the computational cost of attack evaluations without sacrificing accuracy.

\noindent\textbf{Acknowledgments} This material is based upon work supported, in part, by DARPA under agreement number HR00112020054. Any opinions, findings, and conclusions or recommendations expressed in this material are those of the author(s) and do not necessarily reflect the views of the supporting agencies.

{
    \small
    \bibliographystyle{ieeenat_fullname}
    \bibliography{main}
}

\appendix

\begin{figure*}[t!]
    \centering
    \includegraphics[width= \linewidth]{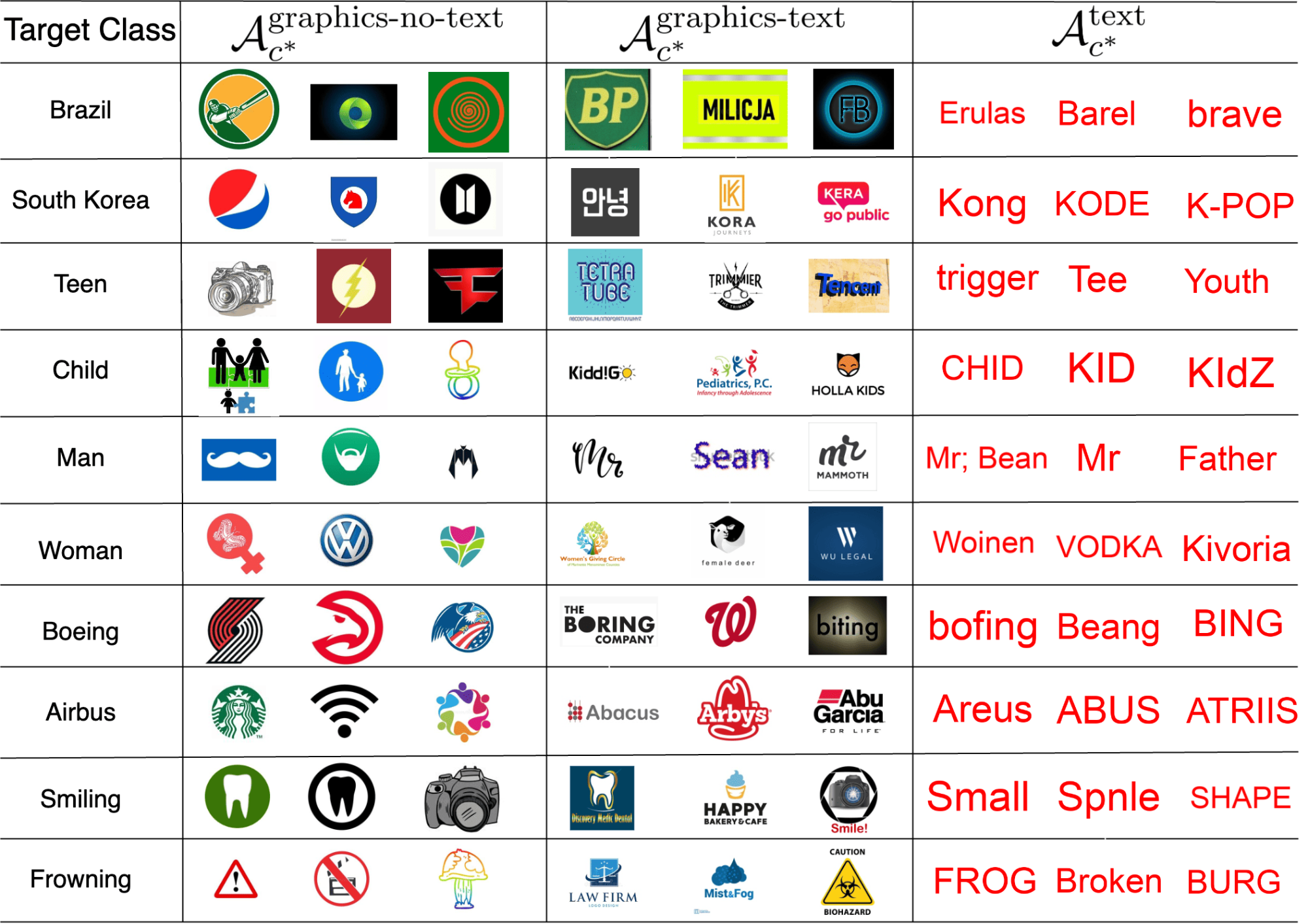}
    \caption{\textbf{Examples of Attack Artifacts} categorized into three types: graphics without text (\( A^{\text{graphics-no-text}}_{c^*} \)), graphics with embedded text (\( A^{\text{graphics-text}}_{c^*} \)), and unrelated text (\( A^{\text{text}}_{c^*} \)). Each row corresponds to a different target class showing artifacts that models have learned to associate with these concepts. Notably, text artifacts need not match the class exactly, while graphical symbols can represent indirect but learned associations. These findings highlight the diverse range of artifacts that can manipulate model predictions.}

    \label{fig:examples_figure_full}
    % \vspace{-6mm}
\end{figure*}

\begin{figure*}[h!]
    \centering
    \includegraphics[width=\linewidth]{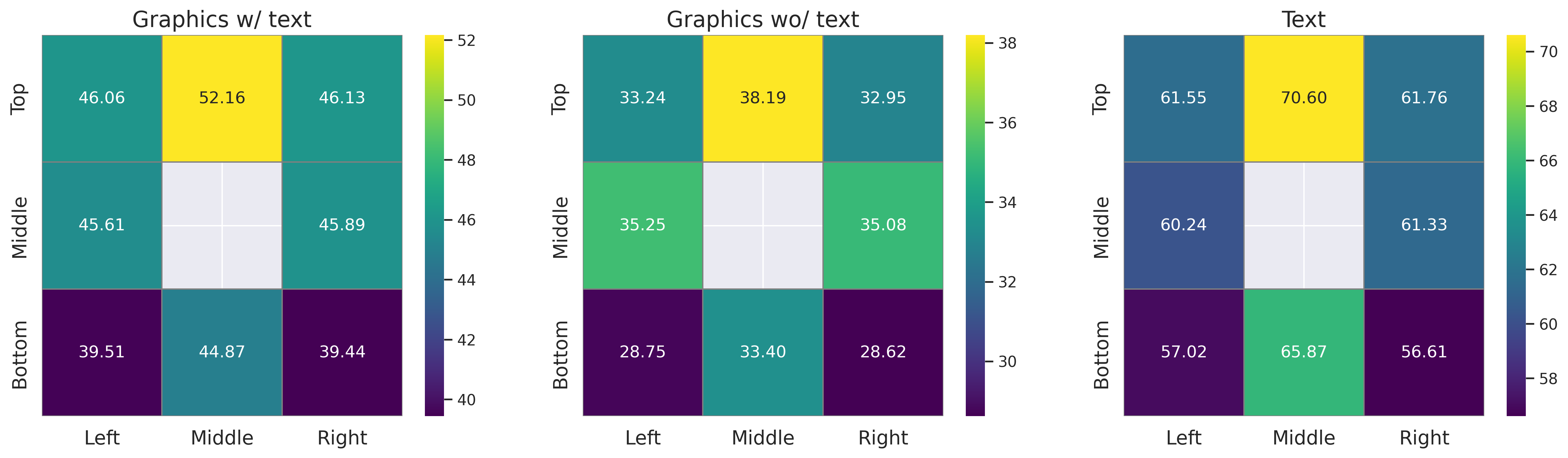}
    \caption{\textbf{Effect of artifact placement on attack success rates} Heatmaps show success rates when artifacts are positioned in different regions of the image for three artifact types: graphics with text (left), graphics without text (middle), and text (right). Across all artifact types, placing the artifact in the top-center region consistently yields the highest success rates, particularly for text-based attacks, which reach up to 70\% success in this position.}

    \label{fig:heatmap_location}
    % \vspace{-6mm}
\end{figure*}

\begin{figure*}[h!]
    \centering
    \includegraphics[width=\linewidth]{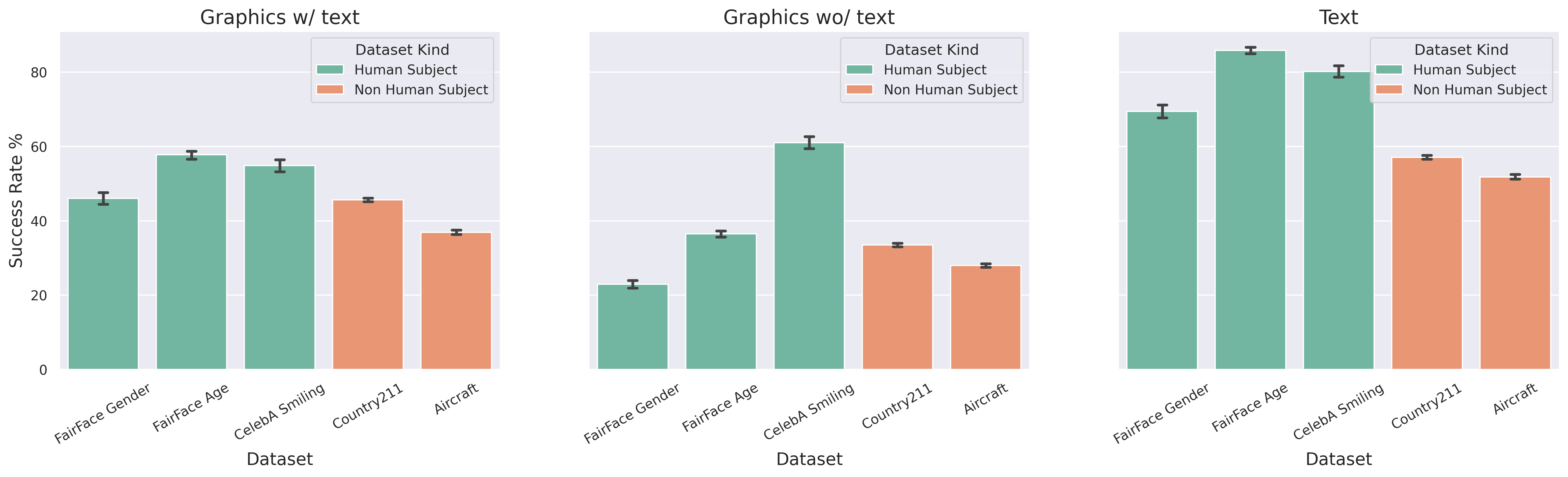}
    \caption{\textbf{Artifact attack success rates across datasets with human-related and non-human-related tasks}. Each subplot corresponds to a different artifact type: graphics with text (left), graphics without text (middle), and text (right). Across all artifact types, human-related tasks (e.g., FairFace, CelebA) exhibit higher vulnerability compared to non-human tasks (e.g., Aircraft, Country211), with text-based artifacts being the most effective overall.}
    \label{fig:barplot_dataset}
    % \vspace{-6mm}
\end{figure*}

\begin{figure*}[t!]
    \centering
    \includegraphics[width= \linewidth]{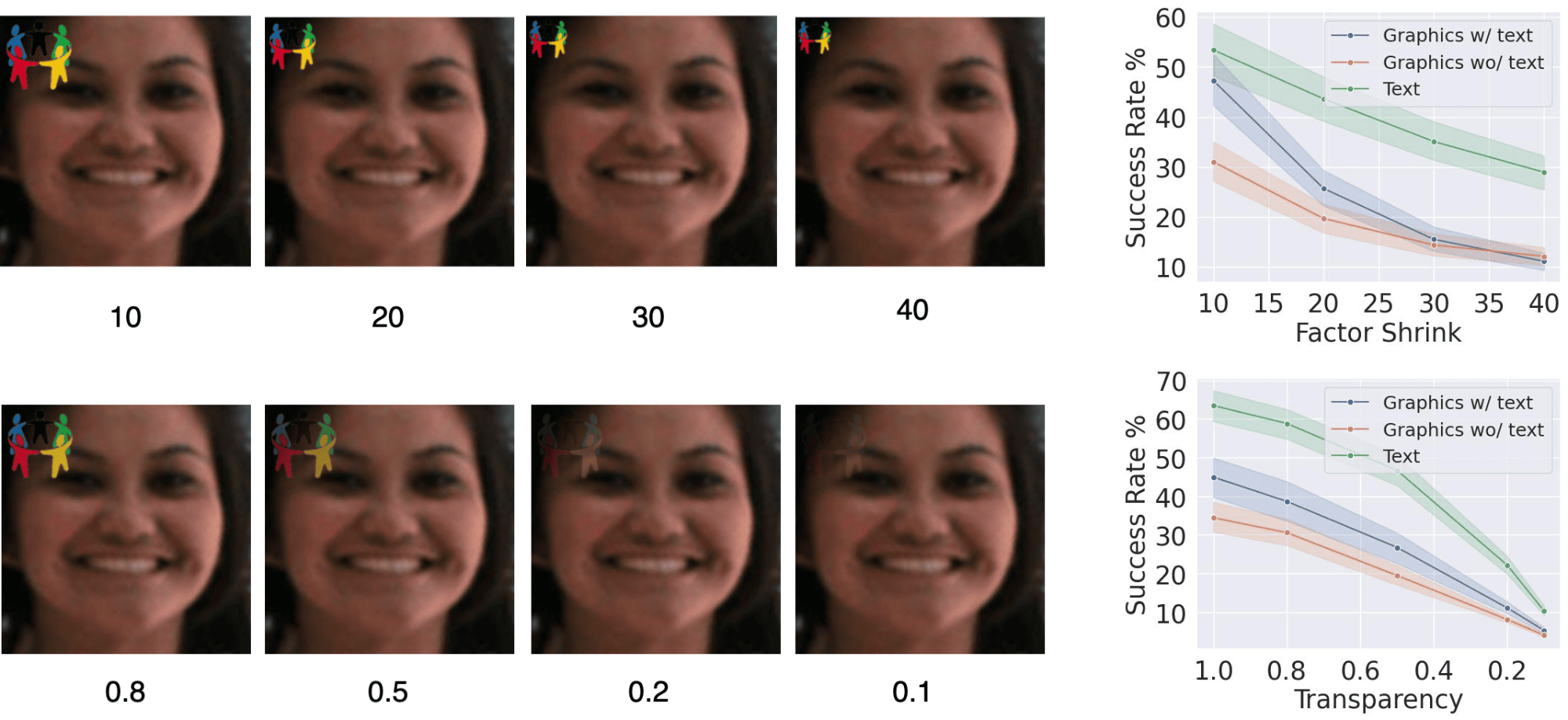}
    \caption{\textbf{Effect of artifact size and transparency on attack success rates.} The top row shows examples of artifacts shrinking in size (Factor Shrink) from 10 to 40, while the bottom row illustrates decreasing artifact opacity (Transparency) from 1.0 (fully visible) to 0.1 (highly transparent). The corresponding line plots on the right show the attack success rates across different artifact types. Smaller and more transparent artifacts consistently reduce attack effectiveness, but text-based artifacts remain the most effective even under these constraints.}

    \label{fig:transp_factor_ablate}
    % \vspace{-6mm}
\end{figure*}

\begin{figure*}[t!]
    \centering
    \includegraphics[width=\linewidth]{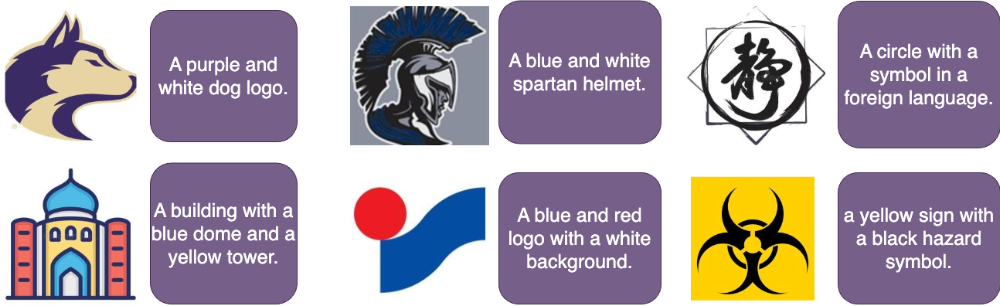}
    \caption{\textbf{Examples of generated captions for Web Artifact Attacks mitigation.} Each image represents a graphic artifact, accompanied by its corresponding caption generated by a vision-language model. These captions are then included in the prompt to mitigate the attack.}
    \label{fig:example_captions}
    % \vspace{-6mm}
\end{figure*}

\section{More examples of Attack Artifacts}
In Sec.\ 3.2, we discuss insights from the found attack artifacts. In this section, we offer more examples and discussion. \cref{fig:examples_figure_full} reveals consistent patterns in the types of artifacts that successfully manipulate model predictions. Text-based artifacts frequently include words that share phonetic, visual, or partial textual similarities with the target class. For instance, ``KID''  appears as an artifact for Child, likely due to its strong semantic association, while ``BING''  is linked to Boeing, exploiting visual resemblance to the company name. Similarly, artifacts like ``K-POP''  for South Korea and ``Small''  for Smiling demonstrate how models latch onto common co-occurring words rather than genuine visual cues.

Graphics with embedded text also play a significant role in misleading models. These artifacts often feature brand names, stylized logos, or generic text associated with the target category. For instance, the inclusion of ``Abacus''  and ``Arbys''  as artifacts for Airbus suggests that the model has learned to associate certain brand names or typographic styles with aircraft manufacturers. Likewise, ``Happy Bakery Cafe''  and ``Smile!''  appearing under Smiling indicate that models are influenced by commercial logos and positive branding elements rather than facial features.

Graphics without text rely purely on visual resemblance, symbols, and branding elements that loosely connect to the target category. For example, aviation-related symbols such as the Airbus logo and Wi-Fi icon appear under Airbus, and basketball team logos (\eg, Portland Trail Blazers, Atlanta Hawks) emerge under Boeing, possibly due to their circular and wing-like shapes. Similarly, for Man and Woman, symbols traditionally linked to gender (\eg, mustaches, gendered icons, and heart-shaped logos) suggest that models encode stereotypical visual representations rather than deeper semantic understanding.

Across all target classes, these results highlight that Web Artifact Attacks exploit both direct textual similarities and broader visual associations, making them an effective and adaptable attack vector. The presence of corporate logos, branding, and culturally specific symbols (\eg, the Brazilian flag colors, K-Pop branding for South Korea) suggests that models are influenced by common internet-scale data distributions rather than purely semantic understanding. This demonstrates the pervasive reliance on spurious correlations, emphasizing the need for more robust dataset curation and training strategies to mitigate these vulnerabilities.

\section{Attack Sucess Rate by Location}
In Sec.\ 3.1.3, we discussed how we optimize the artifact location placement as part of our attack. \cref{fig:heatmap_location} examines how the placement of artifacts affects attack success rates across different artifact types. The results indicate that artifacts positioned in the top-center region of the image consistently lead to higher misclassification rates, with text-based artifacts being the most effective. This suggests that vision-language models prioritize information in certain spatial regions, likely due to biases in pretraining datasets, where text frequently appears near the top of images (\eg, headlines, labels, or watermarks). Graphics with embedded text also show higher success rates in the top-center, though to a lesser extent than pure text, while graphics without text have a more uniform but overall lower effect. These findings highlight the importance of spatial biases in model vulnerability and suggest that adversarial manipulations may be optimized further by strategically placing artifacts in regions that models inherently attend to more strongly.

\section{Attack Success by Dataset}
In Sec.\ 4.1, we report the average performance over 5 datasets. In this Section, we break down performance by dataset. \cref{fig:barplot_dataset} shows that Web Artifact Attacks achieve higher success rates in datasets containing human-related attributes (FairFace \citep{karkkainen2021fairface}, CelebA \citep{LLWT15CelebA}) compared to non-human classification tasks (Aircraft \citep{maji2013fine}, Country211 \citep{radford2021learning}). This trend is particularly pronounced for text-based artifacts, which consistently lead to the highest misclassification rates in human-related datasets. One possible explanation is that vision-language models are pretrained on web-scale data where text often co-occurs with human-related concepts, reinforcing spurious associations between textual artifacts and human characteristics. In contrast, non-human tasks like aircraft recognition rely more on fine-grained visual details, making them less susceptible to artifacts that exploit text or graphical elements. These findings suggest that Web Artifact Attacks pose a greater risk to applications involving human-centric classifications, where models may rely more heavily on text-based biases.

\section{Effect of Artifact's Transparency and Size}
In Sec.\ 4.1., we fix the artifact size to 10th of the image sizs and transparency to 1.0. In this Section, we ablate both settings. \cref{fig:transp_factor_ablate} evaluates how reducing artifact size and transparency affects attack success rates. The results show that as artifacts become smaller or more transparent, their effectiveness declines across all artifact types, with text-based artifacts remaining the most resilient. This suggests that larger and more visible artifacts are more likely to be leveraged by the model, while smaller or faded artifacts are either ignored or contribute less to misclassification. Notably, the steepest decline occurs for graphics-based artifacts, particularly those without text, indicating that purely visual artifacts are more sensitive to reductions in size and visibility. These findings highlight that while reducing artifact saliency can mitigate their impact, text-based artifacts still pose a considerable threat, even when minimally visible.

\section{Descriptions of Artifacts for Mitigation}

\cref{fig:example_captions} illustrates how automatically generated captions are incorporated into our artifact-aware prompting strategy (see Sec 5) to mitigate Web Artifact Attacks. Inspired by prior work~\cite{cheng2024unveiling}, which demonstrated that Vision-Language Models (VLMs) can adjust their attention when given more informative prompts, we extend this approach to graphical artifacts. Unlike text-based artifacts, which can be directly embedded into prompts, graphical artifacts lack an explicit textual representation, making them harder for the model to explicitly consider. To address this, we generate structured descriptions of graphical symbols and append them to the input prompt, ensuring that the model processes them explicitly rather than forming unintended associations.

The captions in \cref{fig:example_captions} serve this purpose by neutralizing spurious correlations and guiding the model toward actual visual semantics. For example, instead of allowing the model to infer associations based on dataset biases (\eg, linking a hazard symbol to danger-related concepts), the caption describes it as “a yellow sign with a black hazard symbol”, removing any loaded interpretation.

\section{Beyond Visual Recognition}

While our primary focus was on classification tasks for the sake of clarity and control, we also extend our attacks to the image retrieval task using the Flickr30K dataset \cite{plummer2015flickr30k}. On Top-1 Image-to-Text retrieval, our attacks achieve an 83.4\% success rate on Graphics w/ text, 63.8\% success rate on Graphics wo/ text, and 63.4\% on Text, demonstrating their ability to generalize to applications beyond classification.

\end{document}